\documentclass{article}

\usepackage{PRIMEarxiv}
\usepackage[utf8]{inputenc} 
\usepackage[T1]{fontenc}    
\usepackage{hyperref}       
\usepackage{url}            
\usepackage{booktabs}       
\usepackage{amsfonts}       
\usepackage{nicefrac}       
\usepackage{microtype}      
\usepackage{lipsum}
\usepackage{fancyhdr}       
\usepackage{graphicx}       
\graphicspath{{media/}}     
\usepackage{subcaption}
\usepackage{tabularx}
\usepackage{booktabs}
\usepackage{url}
\usepackage{enumitem}
\usepackage{listings}
\usepackage{xcolor}

\newcommand{\pytopicgram}{\texttt{pytopicgram}}

\lstdefinestyle{pythonstyle}{
    language=Python,
    basicstyle=\ttfamily\scriptsize,
    keywordstyle=\color{blue},
    commentstyle=\color{gray},
    stringstyle=\color{orange},
    numbers=left,
    numberstyle=\tiny\color{gray},
    frame=single,
    breaklines=true,
    tabsize=4,
    showstringspaces=false
}

\lstdefinestyle{bashstyle}{
    language=Bash,
    basicstyle=\ttfamily\scriptsize,
    keywordstyle=\color{red},
    commentstyle=\color{gray},
    stringstyle=\color{orange},
    numbers=left,
    numberstyle=\tiny\color{gray},
    frame=single,
    breaklines=true,
    tabsize=4,
    showstringspaces=false
}

\title{\pytopicgram: A library for data extraction and topic modeling from Telegram channels}

\author{  
  J. Gómez-Romero, J. Cantón Correa, R. Pérez Mercado, F. Prados Abad, M. Molina-Solana, W. Fajardo \\
  Department of Computer Science and Artificial Intelligence \\
  Universidad de Granada, Spain \\
}

\begin{document}
\maketitle

\begin{abstract}
Telegram is a popular platform for public communication, generating large amounts of messages through its channels. \pytopicgram\ is a Python library that helps researchers collect, organize, and analyze these Telegram messages. The library offers key features such as easy message retrieval, detailed channel information, engagement metrics, and topic identification using advanced modeling techniques. By simplifying data extraction and analysis, \pytopicgram\ allows users to understand how content spreads and how audiences interact on Telegram. This paper describes the design, main features, and practical uses of \pytopicgram, showcasing its effectiveness for studying public conversations on Telegram.
\end{abstract}


\section{Motivation and significance}

Messaging platforms like Telegram have become critical spaces for information exchange, social mobilization, and digital communities. With features such as public channels, unlimited subscribers, and a degree of anonymity, Telegram has emerged as a valuable source of unstructured data reflecting various social, political, and cultural dynamics \cite{Duportail2020}. Understanding the discourses within Telegram channels is essential for researchers, policymakers, and organizations seeking to analyze public opinion, track information diffusion, and monitor emerging trends, including harmful content like disinformation.

Topic modeling is an unsupervised learning technique that identifies patterns in text by grouping words into topics. Traditional methods, such as Latent Dirichlet Allocation (LDA) \cite{Blei2003}, have been improved by modern approaches that use Large Language Models (LLMs) like BERT and GPT to capture more context-aware representations of text \cite{Alammar2024}. These techniques leverage the deep contextual embeddings provided by the LLMs and enable more accurate and semantically-rich topic representations.

To address the need for analyzing Telegram data, we present \pytopicgram\footnote{\url{https://github.com/ugr-sail/pytopicgram}}, a Python library for downloading, processing, and categorizing messages from Telegram channels. The library relies on the Telethon\footnote{\url{https://docs.telethon.dev/en/stable/}} library to connect to the Telegram API for data collection and the BERTopic algorithm\footnote{\url{https://github.com/MaartenGr/BERTopic}} for topic modeling. These features make \pytopicgram\ particularly effective for analyzing public discourse, where understanding the spread and evolution of narratives is crucial.

This paper describes \pytopicgram’s architecture, functionalities, and impact. Two use examples are also provided to illustrate running the complete toolchain and selectively using one of the modules to perform recursive data collection.

\section{Software description}

\subsection{Software architecture}
\begin{figure*}[t]
\centering
\includegraphics[width=0.95\textwidth]{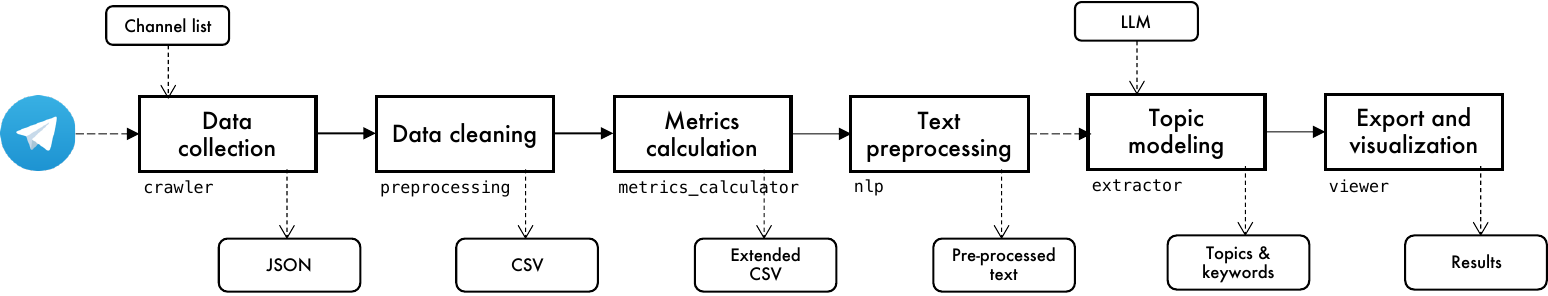}
\caption{Architecture of \pytopicgram}\label{fig1:architecture}
\end{figure*}

\pytopicgram\ integrates several tools and libraries to provide a streamlined pipeline for message crawling, preprocessing, metrics calculation, natural language processing, and topic modeling. The architecture is organized into key modules, each responsible for a specific task in the data analysis pipeline (Figure \ref{fig1:architecture}). The \texttt{main.py} script orchestrates the entire pipeline, where each module produces output files utilized by subsequent modules. Additionally, the modules can run in stand-alone mode, offering flexibility depending on the use case.

\begin{itemize}
\item \textbf{Data collection:} This module uses the Telethon library to connect to the Telegram API and retrieve messages from specified public channels. Users can specify channel names, timeframes, and apply filters. The collected data includes message text, along with metadata such as timestamps, views, and forwards, which are saved in JSON format for later processing.

\item \textbf{Data cleaning:} This module prepares raw Telegram message data for analysis by transforming and cleaning them. JSON data is converted into a data frame, allowing for feature selection from user-defined lists, and elements such as URLs, emojis, reactions, and mentions are extracted. The processed data is saved in CSV format, ready for the next steps in the pipeline.

\item \textbf{Metrics calculation:} This module computes various metrics on the cleaned metadata, quantifying different aspects of the messages. Metrics like the virality ratio \cite{Nobari2021} are calculated, and a simplified media type is assigned to each message. The output, a CSV file, includes the preprocessed data alongside the calculated metrics.

\item \textbf{Text preprocessing:} This module applies NLP techniques, including sentence splitting and polarity/subjectivity calculation, to the text of the cleaned messages. Python libraries such as spaCy and NLTK are employed for efficient text processing. The output, containing processed text data with additional NLP features, is saved as a CSV file for use in the next stage.

\item \textbf{Topic modeling:} This core module uses the BERTopic algorithm to perform topic modeling and extract relevant topics from the text data. It leverages embeddings from Large Language Models (LLMs) to create context-aware topic representations. Additionally, this module can integrate with OpenAI’s GPT to generate topic descriptions. The output includes the topic model and details about the messages used for training, saved in model files.

\item \textbf{Export and visualization:} This module provides functionality for exporting analyzed data and models in various formats and visualizing the results of the topic modeling process. Visualizations, such as the evolution of topics over time, hierarchical clustering, and relationships between topics and keywords, are generated and can be saved in different formats for reporting or presentation purposes.
\end{itemize}

\pytopicgram\ is built in Python and requires Python 3.7 or higher. All dependencies are installed via Python’s package manager, \texttt{pip}. Access to the Telegram API and the necessary API keys are required for data collection, and an OpenAI key is required for generating topic descriptions using GPT.

\subsection{Software functionalities}
\pytopicgram\ offers a powerful set of functionalities for analyzing Telegram messages:

\begin{itemize}

\item \textbf{Fast and flexible message crawling:} \pytopicgram\ uses the Telethon library to efficiently connect to the Telegram API and retrieve messages from public channels. It supports fast, unlimited crawling, extracting all available data provided by the API, including message text, timestamps, views, forwards, and reactions.

\item \textbf{Extended channel information retrieval:} In addition to message crawling, \pytopicgram\ gathers detailed information about the Telegram channels, including the number of subscribers, channel creation date, recommended channels, profile photo, pinned messages, and more, providing valuable context to the analysis.

\item \textbf{Computation of message metrics and extraction of relevant elements:} \pytopicgram\ calculates various metrics, such as virality ratios, offering insights into content reach and engagement. It also extracts key elements from messages, including URLs, link domains, emojis, and mentions, and tags the media contents of the messages from a combination of metadata.

\item \textbf{Out-of-the-box BERTopic integration:} \pytopicgram\ integrates the BERTopic algorithm seamlessly, enabling users to perform topic modeling without extra configuration. The process is customizable, allowing for adjustments such as the number of initial topics and topic representations.

\item \textbf{Language-agnostic capabilities:} Although the use cases in this paper revolve around Spanish-language content, \pytopicgram\ is fully capable of handling multiple languages, making it versatile for a wide range of linguistic contexts.

\item \textbf{Data minimization and process optimization:} \pytopicgram\ allows users to limit the message features stored at the beginning of the analysis, supporting data minimization and reducing dataset size. Additionally, the number of texts used for model training can be controlled, optimizing processing while still allowing classification of larger datasets once the model is trained.

\end{itemize}

\section{Illustrative Examples}

\subsection{Example 1: Running the complete pipeline from command line}

In this example, we demonstrate the full functionality of \pytopicgram\ by running the entire pipeline through the \texttt{main.py} script. The following command initiates the process, where messages are downloaded, preprocessed, and analyzed using topic modeling:

\begin{lstlisting}[style=bashstyle]
python main.py 
    --api_id <TELEGRAM_API_ID> 
    --api_hash <TELEGRAM_API_HASH> 
    --start_date 2024-08-01T00:00:00+00:00  
    --end_date   2024-09-01T00:00:00+00:00 
    --channels_file config/channels_sample.csv 
    --openai_key <OPENAI_KEY> 
    --description "Sample running, Aug 2024, using OpenAI API"
\end{lstlisting}

The main parameters of \texttt{main.py} are the following:
\begin{itemize}
    \item \texttt{api\_id}: The API ID obtained from the Telegram Developer Portal.
    \item \texttt{api\_hash}: The API hash corresponding to the API ID.
    \item \texttt{start\_date 2024-08-01T00:00:00+00:00}: Specifies the start date for crawling messages, in this case, August 1, 2024.
    \item \texttt{end\_date 2024-09-01T00:00:00+00:00}: Specifies the end date for crawling messages, in this case, September 1, 2024.
    \item \texttt{channels\_file config/channels\_sample.csv}: The CSV file containing the list of Telegram channels from which messages will be crawled.
    \item \texttt{openai\_key}: An OpenAI API key used to generate topic descriptions through GPT. 
    \item \texttt{description "Sample running, Aug 2024, using OpenAI API"}: A descriptive text that provides context for the current pipeline run. This description is logged and saved with the output files to ensure traceability of the analysis process.
\end{itemize}

In addition to the core arguments used in this example, \pytopicgram\ provides several other useful parameters for enhanced flexibility. For instance, users can limit the number of messages processed using the \texttt{limit} option, or extract specific elements from the text, such as URLs, emojis, and mentions, through the \texttt{capture\_urls}, \texttt{capture\_emojis}, and \texttt{capture\_mentions} flags. The calculation of the metrics can also be parametrized, e.g., the number of neighbors to calculate virality and the threshold to consider a mesage viral. For topic modeling, the \texttt{extractor\_sample\_ratio} parameter allows building the model from a sample of messages, which can be useful when dealing with large datasets. Additionally, the \texttt{viewer\_generate\_viz} option enables generation of visualizations for inital exploration of the results.

Upon execution, the pipeline generates several output files. The main ones are: the annotated messages file (by default \texttt{messages\_annotated.csv}), which contains the messages with metadata, metrics and their assigned topics; the topic information file (\texttt{topic\_info.csv}), which provides detailed information about each topic, including keywords and various topic representations; and the channel extended information file (\texttt{channels\_list\_details.csv}), which includes channel data such as recommended channels and subscriber counts. Additionally, a JSON file with the data provided by Telegram API, a BERTopic model file in pickle format, and a few HTML files with visualizations are generated for further analysis and reuse. 

\subsection{Example 2: Snowball channels' data collection}
This example demonstrates the flexibility of \pytopicgram's crawler functions \texttt{read\_channels\_from\_csv} and \texttt{process\_channels}. We began by downloading messages from a set of channels listed in a CSV file. After processing the initial set of channels, we utilized the recommended channels from each to continue the process recursively, effectively ``creating a snowball'' through the Telegram network. The output is a progressively expanding dataset of Telegram messages, accompanied by a CSV file containing the full list of channels processed. Please note that the provided code could be easily extended to perform topic modeling for the intermediate and the final datasets via the \texttt{train\_bertopic\_model} function of the extractor.

\begin{lstlisting}[style=pythonstyle]
# Read and initialize the initial set of channels    
current_channels = read_channels_from_csv(initial_csv)
current_channels['similar_channels'] = 
    [[] for _ in range(len(current_channels))]
current_channels.to_csv(temp_path, index=False, 
    mode='w', header=True)
all_channels = pd.DataFrame()  
visited_channels_info = pd.DataFrame()  
round_counter = 1

while round_counter <= max_rounds and not current_channels.empty:
    print(f"\n[bold cyan]Snowball Round {round_counter}/{max_rounds}")
        
    # Run the process_channels function to collect messages and channel info
    print(f"[cyan]Processing {len(current_channels)} channels in round {round_counter}...")
    current_channels.reset_index(drop=True, inplace=True)
    await process_channels(current_channels, start_date, end_date, api_id, api_hash, messages_csv, channels_file_name=temp_filename, append=(round_counter > 1), by_url=(round_counter == 1))
        
    # Load found channels information
    current_channels = pd.read_csv(temp_path, index_col=False)

    # Collect new channels recommended by the current set
    new_channels_list = []
    for _, row in current_channels.iterrows():
        similar_channels = row.get('similar_channels', [])
        similar_channels_list = eval(similar_channels.encode('utf-8').decode('unicode_escape'))
        for recommended_channel in similar_channels_list: 
            new_channels_list.append({
                'id': recommended_channel['id'],
                'channel_name': recommended_channel['title'], 
                'url': f"https://t.me/{recommended_channel['username']}", 
                'cluster': "not assigned",
                'user': recommended_channel['username']
            })
        
    # Convert to DataFrame for the next round
    new_channels_df = pd.DataFrame(new_channels_list).drop_duplicates(subset='id')

    # Add current channels to the cumulative dataset
    all_channels = pd.concat([all_channels, current_channels], ignore_index=True)

    # Keep the info of visited channels
    visited_channels_info = pd.concat([visited_channels_info, current_channels], ignore_index=True)

    # Remove channels that were already processed
    new_channels_df = new_channels_df[~new_channels_df['id'].isin(all_channels['id'])]

    # Update the channels for the next round
    current_channels = new_channels_df
    round_counter += 1

    if new_channels_df.empty:
        print("[green]No new channels found. Snowball complete.")
        break
\end{lstlisting}

\section{Impact}

Telegram's increasing role in public communication, particularly through its public channels, has made it an important platform for monitoring various aspects of public discourse. Researchers can use Telegram data to explore a wide range of topics, such as political movements, cultural trends, market behaviors, and grassroots initiatives. Channels on Telegram provide a rich source of information on how communities mobilize, share knowledge, and engage in discussions, making the platform valuable for understanding trends in public opinion and the spread of ideas. These applications highlight Telegram's broader relevance for tracking societal shifts and emerging phenomena across different fields.

In addition to these broader uses, Telegram's open and minimally moderated environment makes it a significant platform for the dissemination of disinformation, hate speech, and harmful narratives. The rapid spread of false or manipulative content on Telegram channels, often with little oversight, poses challenges for researchers, policymakers, and civil society groups. Monitoring such content is crucial for understanding how narratives evolve, how they are shared, and how they influence public sentiment, particularly in critical areas such as political disinformation, health misinformation, and extremist rhetoric.

\pytopicgram\ addresses the need for analyzing large volumes of Telegram data by providing an integrated solution that combines message crawling, preprocessing, and unsupervised topic modeling. The use of an unsupervised approach means that \pytopicgram\ does not require pre-labeled data or annotations, which, while potentially less precise than supervised methods, allows for greater scalability across diverse datasets and topics. This makes it especially valuable for exploratory analysis when dealing with vast or evolving data sets.

Compared to directly using Telegram’s API and BERTopic libraries, \pytopicgram\ automates the key steps of data collection and processing, allowing users to perform the entire analysis pipeline without the need for separate configuration or coding. This streamlines the workflow significantly and ensures that all steps are integrated into a coherent process, reducing the risk of inconsistencies in data handling.

Compared to other tools like TGstat\footnote{\url{https://tgstat.com}}, which provides valuable channel statistics and insights but relies on external services to gather and present data, \pytopicgram\ offers the advantage of enabling users to collect, store, and analyze Telegram data locally. This ensures privacy, complete data ownership, and the ability to configure the entire analysis programmatically, which is crucial for custom research needs and deeper analyses. In contrast to Telegram Tracker\footnote{\url{https://github.com/congosto/telegram-tracker-t-hoarder_tg}}, which focuses on message extraction and basic metrics, \pytopicgram\ integrates advanced natural language processing techniques to extract relevant pieces of data (e.g., links and mentions), and perform contextually aware topic modeling. The combination of topic modeling with crawling and preprocessing capabilities offers a more comprehensive solution than simpler extraction-focused alternatives.

The ability to analyze message content in relation to engagement metrics, such as views, forwards, reactions, comments, and calculated virality ratios, makes \pytopicgram\ well-suited for studying the dissemination and impact of messages across Telegram channels. By examining how content spreads and measuring the reach and virality of specific messages, as well as user engagement through reactions and comments, the tool offers valuable insights into which narratives resonate with audiences and how they evolve over time. Beyond engagement tracking, \pytopicgram's metrics support broader applications in content monitoring, enabling trend detection, topic summarization, and the identification of influential messages, making it a versatile solution for media content analysis.

The ability to analyze message content in relation to engagement metrics, such as views, forwards, reactions, comments, and calculated virality ratios, makes \pytopicgram\ well-suited for studying the dissemination and impact of messages across Telegram channels. By examining how topics spread and measuring the reach and virality of messages, the tool provides insights into which narratives resonate with audiences and how they evolve. In addition to tracking propagation, engagement, and narratives, \pytopicgram\ extracts extended channel data, including related channels, description text, and pinned messages, offering a comprehensive view of channel activity. These features can be used individually or combined, providing a flexible toolkit for diverse content analysis needs.

Given Telegram's significance in shaping public discourse, tools like \pytopicgram\ are essential for systematically examining content and supporting evidence-based research. Its functionality aligns with the growing need for monitoring large platforms, especially in light of regulatory frameworks like the European Union's Digital Services Act \cite{EuropeanCommission2020}, which aims to ensure safer digital environments.

\section{Conclusions}
\pytopicgram\ offers an integrated solution for collecting, processing, and analyzing large volumes of Telegram data. Its unsupervised topic modeling approach, combined with detailed engagement metrics and extended channel information, provides a comprehensive view of the narratives that unfold across Telegram channels. This makes it a versatile tool for a wide range of media content analysis tasks, from summarizing channel discussions to studying content virality and influence. Compared to other tools, \pytopicgram\ stands out for its flexibility, local data processing capabilities, and its ability to scale for large datasets without requiring annotated data. As Telegram continues to be a significant platform for public discourse, tools like \pytopicgram\ will play a crucial role in enabling researchers and analysts to explore the evolving landscape of online communication.

\section*{CRediT authorship contribution statement}
J. Gómez-Romero: Conceptualization, Investigation, Software, Writing – original draft. J. Cantón Correa: Conceptualization, Investigation, Writing – review \& editing. R. Pérez Mercado: Investigation, Software, Writing – review \& editing. F. Prados Abad: Investigation, Software, Writing – review \& editing. M. Molina-Solana: Investigation, Software, Writing – review \& editing. W. Fajardo: Investigation, Software, Writing – review \& editing.


\section*{Data availability}
We have shared the code repository and the link is provided in the paper.

\section*{Acknowledgments}
This work was supported by the UDDOT project funded by European Media Information Fund (EMIF) managed by the Calouste Gulbenkian Foundation and the XAI-DISINFODEMICS project (PLEC2021-007681) funded by MICIU/AEI/10.13039/501100011033 and by European Union NextGenerationEU/PRTR.
\bibliographystyle{elsarticle-num}  
\bibliography{cas-refs} 

\end{document}